\documentclass[conference]{lib/IEEEtran}
\IEEEoverridecommandlockouts
\usepackage{cite}
\usepackage{amsmath,amssymb,amsfonts}
\usepackage{algorithmic}
\usepackage{graphicx}
\usepackage{textcomp}
\usepackage{xcolor}
\usepackage{lipsum}
\usepackage{booktabs, multirow}
\usepackage[utf8]{inputenc}
\usepackage[T1]{fontenc}

\usepackage[font={small}]{caption}
\usepackage[backref]{hyperref}
\usepackage{cite}

\usepackage{todonotes}
\def\BibTeX{{\rm B\kern-.05em{\sc i\kern-.025em b}\kern-.08em
    T\kern-.1667em\lower.7ex\hbox{E}\kern-.125emX}}

\usepackage{url}

\newcommand{\etal}{\emph{et al.}}
\newcommand{\sota}{state-of-the-art }

\begin{document}
    \title{
    Skip-GANomaly: Skip Connected and Adversarially Trained Encoder-Decoder Anomaly Detection
}

\author{
    \IEEEauthorblockN{Samet Ak\c{c}ay, Amir Atapour-Abarghouei, Toby P. Breckon}
    \IEEEauthorblockA{
        \textit{Department of Computer Science, Durham University}, Durham, UK \\
        {\tt\small
        \{\href{mailto:samet.akcay@durham.ac.uk}
               {samet.akcay},
          \href{mailto:amir.atapour-abarghouei@durham.ac.uk}    
               {amir.atapour-abarghouei},
          \href{mailto:toby.breckon@durham.ac.uk}
               {toby.breckon}\}@durham.ac.uk
        }
    }
}

\maketitle

\begin{abstract}

   Despite inherent ill-definition, anomaly detection is a research endeavour of great interest within machine learning and visual scene understanding alike. Most commonly, anomaly detection is considered as the detection of outliers within a given data distribution based on some measure of normality. The most significant challenge in real-world anomaly detection problems is that available data is highly imbalanced towards normality (i.e. non-anomalous) and contains a most a sub-set of all possible anomalous samples - hence limiting the use of well-established supervised learning methods. By contrast, we introduce an unsupervised anomaly detection model, trained only on the normal (non-anomalous, plentiful) samples in order to learn the normality distribution of the domain, and hence detect abnormality based on deviation from this model. Our proposed approach employs an encoder-decoder convolutional neural network with skip connections to thoroughly capture the multi-scale distribution of the normal data distribution in high-dimensional image space. Furthermore, utilizing an adversarial training scheme for this chosen architecture provides superior reconstruction both within high-dimensional image space and a lower-dimensional latent vector space encoding. Minimizing the reconstruction error metric within both the image and hidden vector spaces during training aids the model to learn the distribution of normality as required. Higher reconstruction metrics during subsequent test and deployment are thus indicative of a deviation from this normal distribution, hence indicative of an anomaly. Experimentation over established anomaly detection benchmarks and challenging real-world datasets, within the context of X-ray security screening, shows the unique promise of such a proposed approach. 

\end{abstract}

    \section{Introduction}
\label{sec:introduction}

\begin{figure}
    \centering
    \includegraphics[width=\linewidth]{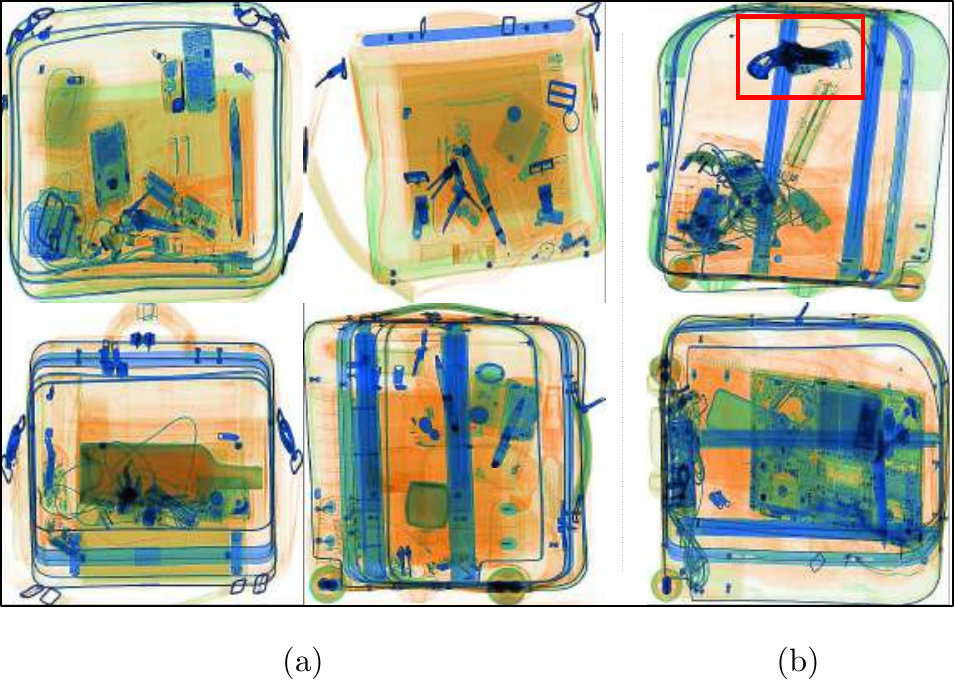}
    \caption{Sub-sample of the X-ray screening application dataset used to train the proposed approach: (a) training data contains normal samples only, while the test data (b) comprises both normal and abnormal samples.}
    \label{fig: introduction}
\end{figure}

Anomaly detection is an increasingly important area within visual image understanding. Following recent trends in the field, there has been a significant increase in the availability of large datasets. However, in most cases such data resources are highly imbalanced towards examples of normality (non-anomalous), whilst lacking in examples of abnormality (anomalous) and offering only partial coverage of all possibilities can could encompass this latter class. This variation, and somewhat unknown nature, of the anomalous class mean such datasets lack the capacity and diversity to train traditional supervised detection approaches. In many application scenarios, such as the X-ray screening example illustrated in Figure \ref{fig: introduction}, the availability of anomalous cases may be limited and may evolve over time due to external factors. Within such scenarios, unsupervised anomaly detection has become instrumental in modeling such data distributions, whereby the model is trained only on normal (non anomalous) samples to capture the distribution of normality, and then evaluated on both unseen normal and abnormal (anomalous) examples to find their deviation from the distribution.

A significant body of prior work exists within anomaly detection for visual scene understanding \cite{Markou2003a,Markou2003,Hodge2004,Chandola2009,Pimentel2014} with a wide range of application domains \cite{Abdallah2016, Ahmed2016, Ahmed2016a, Schlegl2017, Kiran2018}.  A common hypothesis in such anomaly detection approaches is that abnormal samples differ from normality in not only high-dimensional image space but also with lower-dimensional latent space encoding. Hence, mapping high-dimensional images to lower-dimensional latent space becomes essential. The critical issue here is that capturing the distribution of the normal samples is rather challenging. Recent developments in Generative Adversarial Networks (GAN) \cite{Goodfellow2014b}, shown to be highly capable of obtaining input data distribution, have led to a renewed interest in the anomaly detection problem. Several contemporary studies demonstrate that the use of GAN has great promise to address this anomaly detection problem since they are inherently adept at mapping high-dimensional to lower dimensional latent encoding and \textit{vice-versa} with minimal information loss \cite{Schlegl2017, Zenati2018a, Akcay2018b}. 

Schlegl \etal \cite{Schlegl2017} trains a pre-trained GAN backwardly to map from image space to lower-dimensional latent space, hypothesizing that differences in latent space would yield anomalies. Zenati \etal \cite{Zenati2018a} jointly train a two network to capture normal distribution by mapping from image space to latent space, and vice-versa. Ak\c{c}ay \etal \cite{Akcay2018b} trains an encoder-decoder-encoder network with the adversarial scheme to capture the normal distribution within the image and latent space. Sabokrou \etal \cite{Sabokrou2018} also trains an adversarial network to capture the normal distribution, hypothesizing that the model would fail to generate abnormal samples, where the difference between the original and generated images would yield the abnormality. This prior work in the field \cite{Schlegl2017, Zenati2018a, Akcay2018b, Sabokrou2018}, empirically illustrates both the importance and promise of anomaly detection anomalies within dual image and latent space.

Here we propose a new method for anomaly detection via the adversarial training over a skip-connected encoder-decoder (convolutional neural) network architecture. Whilst adversarial training has shown the promise of GAN in this domain \cite{Akcay2018b}, skip-connections within such UNet style (encoder-decoder) \cite{ronneberger2015unet} generator networks are known to enable the multi-scale capture of image space detail with sufficient capacity to generate high-quality normal image drawn from the distribution the model has learned. Similar to \cite{Schlegl2017,Zenati2018a ,Akcay2018b}, the proposed approach also seeks to learn the normal distribution in both the image and latent spaces via a GAN generator-discriminator paradigm. The discriminator network not only forces the generator to learn an improved model of the distribution but also works as a feature extractor such that it learns the reconstruction of the normal distribution within a lower-dimensional latent space. Evaluation of the model on various established benchmarks \cite{cifar10,CAST2016}  statistically illustrate superior anomaly detection task performance over prior work \cite{Schlegl2017,Zenati2018a,Akcay2018b}. Subsequently, the main contributions of this paper are as follow:


\begin{itemize}
    \item \emph{unsupervised anomaly detection} ---  a unique unsupervised adversarial training regime, over a skip-connected encoder-decoder convolutional network architecture, yields superior reconstruction within the image and latent vector spaces.
    \item \emph{efficacy} --- an efficient anomaly detection algorithm achieving quantitatively and qualitatively superior performance against prior state-of-the-art approaches.
    \item \emph{reproducibility} --- a simple yet effective algorithmic approach that can be readily reproduced.
\end{itemize}
\section{Related Work}
\label{sec:related-work}
Anomaly detection is a major area of interest within the field of machine learning with various real-world applications spanning from biomedical\cite{Schlegl2017} to video surveillance\cite{Kiran2018}. Recently, a considerable literature has grown up in the field, leading to a proliferation of taxonomy papers \cite{Markou2003a, Markou2003, Hodge2004, Chandola2009, Pimentel2014}. Due to the current trends, the review in the paper primarily focuses on reconstruction-based anomaly detection approaches.

One of the most influential accounts of anomaly detection using adversarial training comes from Schlegl \etal \cite{Schlegl2017}. The authors hypothesize that the latent vector of the GAN represents the distribution of the data. However, mapping to the vector space of the GAN is not straightforward. To achieve this, the authors first train a generator and discriminator using only normal images. In the next stage, they utilize the pre-trained generator and discriminator by freezing the weights and remap to the latent vector by optimizing the GAN based on the $z$ vector. During inference, the model pinpoints an anomaly by outputting a high anomaly score, reporting significant improvement over the previous work. The main limitation of this work is its computational complexity since the model employs a two-stage approach, and remapping the latent vector is extremely expensive. In a follow-up study, Zenati \etal \cite{Zenati2018a} investigate the use of BiGAN \cite{Donahue2016} in an anomaly detection task, examining joint training to map from image space to latent space simultaneously, and vice-versa. Training the model via \cite{Schlegl2017} yields superior results on the MNIST \cite{LeCun2010} dataset. In a similar study in which image and latent vector spaces are optimized for anomaly detection, Ak\c{c}ay \etal \cite{Akcay2018b} propose an adversarial network such that the generator comprises encoder-decoder-encoder sub-networks. The objective of the model is not only the minimize the distance between the real and fake normal images, but also minimize the distance within their latent vector representations jointly. The proposed approach achieves \sota performance both statistically and computationally.

Taken together, these studies support the notion that the use of reconstruction-based approaches shows promise within the field \cite{Kiran2018, Schlegl2017, Zenati2018a, Akcay2018b, Sabokrou2018}. Motivated by the previous methods in which latent vectors are optimized \cite{Schlegl2017, Zenati2018a, Akcay2018b}, we propose an anomaly detection approach that utilizes adversarial autoencoders with skip connections. The proposed approach learns representations within both image and latent vector space jointly and achieves numerically superior performance. 
\section{Proposed Approach}
\label{sec:our-approach}
Before proceeding to explain our proposed approach, it is important to introduce the fundamental concepts.

\subsection{Background}
\subsubsection{Generative Adversarial Networks (GAN)} \label{ssec:gan}

GAN are unsupervised deep neural architectures that learn to capture any input data distribution by predicting features from an initially hidden representation. Initially proposed in \cite{Goodfellow2014b}, the theory behind GAN is based on a competition of two networks within a zero-sum game framework, as initially used in game theory. The task of the first network, called Generator ($G$) is to capture the distribution of the input dataset for a given class label, by predicting features (\textit{or images}) from a hidden representation, which is commonly a random noise vector. Hence the generator network has a decoder network architecture such that it up-samples the input arbitrary latent representation to generate high dimensional features. The task of the second network, called Discriminator ($D$), on the other hand, is to predict the correct class (i.e., \textit{real vs. fake}) based on the given features (\textit{or images}). The discriminator network usually adopts encoder network architecture such that for a given high dimensional feature, it predicts its class label. With optimization based on a zero-sum game framework, each network strengthens its prediction capability until they reach an equilibrium.

Due to their inherent potential for capturing data distributions, there is a growing body of literature that recognizes the importance of GAN \cite{Creswell2017}. Training two networks jointly to reach an equilibrium, however, is not a straightforward procedure, causing training instability issues. Recently, there has been a surge of interest in addressing the instability issues via several empirical methodologies \cite{Salimans2016, Arjovsky2017a}. An innovative and seminal work of Radford and Chintala \cite{Radford2015} pioneered a new approach to stabilize GAN training by using fully convolutional layers and batch normalization \cite{Ioffe2015} throughout the network. Another well-known attempt to stabilize GAN training is the use of Wasserstein loss in the training objective, which significantly improves the training issues \cite{Arjovsky2017, Gulrajani2017}.

\subsubsection{Adversarial Auto-Encoders (AAE)} 
Conceptually similar to GAN, AAE consist of a generator and a discriminator networks. The generator has a bow-tie architectural network style comprising both an encoder and a decoder. The task of the generator is to reconstruct an input data by down-sampling it into a latent representation first, and then by upsampling the latent vector into the reconstructed data (image). The task of the discriminator network is to predict whether the input is a latent vector from the auto-encoder or the prior distribution initialized arbitrarily. Training AAE provides superior reconstruction as well as the capability of controlling the latent space \cite{Mirza2014, Makhzani2015, Creswell2017}.

\subsubsection{Inference within GAN} 
A strong correlation has been demonstrated between the manipulation of the input noise vector and the output of the generator network \cite{Radford2015, Chen2016}. Similar latent space variables have demonstrably produced visually similar high-dimensional images \cite{Creswell2016}. One approach to finding the optimal latent vectors to create similar images is to inversely map images back to their hidden space via their gradients \cite{Lipton2017}. Alternatively, with an additional encoder network that down-samples high dimensional images into lower dimensional latent space, vanilla GAN are reported to be capable of learning inverse mapping \cite{Donahue2016}. Another way to learn inference via inverse mapping is to jointly train two networks such that the former maps images to latent space, while the latter maps this latent space representation back into higher dimensional image space \cite{Dumoulin2016}. Based on these previous findings, the primary aim of this paper is to explore inference within GAN by exploiting the latent vector representation in order to find unique a representation for a normal (non anomalous) data distribution such that it can be statistically differentiated from unseen, unknown and varying abnormal (anomalous) data samples.

\subsection{Proposed Approach}

\subsubsection{Problem Definition} 
\label{sssec:problem-definition}
This work proposes an unsupervised approach for anomaly detection.



We adversarially train our proposed convolutional network architecture in an unsupervised manner such that the conceptual model is trained on normal samples only, and yet tested on both normal and abnormal ones. Mathematically, we define and formulate our problem as the following:

An input dataset $\mathcal{D}$ is split into train $\mathcal{D}_{trn}$ and test sets $\mathcal{D}_{tst}$ such that $\mathcal{D}_{trn}=\{(x_1, y_1), (x_2, y_2), \ldots,\allowbreak (x_m, y_m)\}$ contains $m$ normal samples, where $y_i=0$ denotes normal class. The test set $\mathcal{D}_{tst} = \{(x_1, y_1), \allowbreak (x_2, y_2), ..., (x_m, y_m)\}$ comprises $n$ normal and abnormal samples, where $y_i \in [0,1]$ for normal and abnormal classes, respectively. In practical setting,  $m \gg n$.

\begin{figure}
    \centering
    \includegraphics[width=\linewidth]{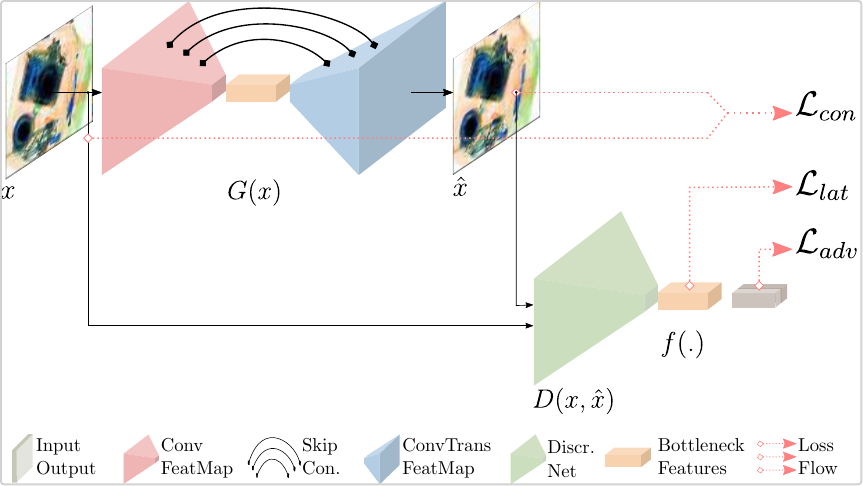}   
    \caption{Overview of the proposed adversarial training procedure.}
    \label{fig:high-level-overview}
\end{figure}

Based on the dataset defined above, we are to train our model $f$ on ${D}_{trn}$ and evaluate its performance on ${D}_{tst}$. The training objective ($\mathcal{J}$) of the model $f$ is to capture the distribution of ${D}_{trn}$ within not only image space but also hidden latent vector space. Capturing the distribution within both dimensions by minimizing $\mathcal{J}$ enable the network to learn higher and lower level features that are unique to normal images. We hypothesize that defining an anomaly score $\mathcal{A}(.)$ based on the training objective $\mathcal{J}$ would yield minimum anomaly scores for training samples ---\textit{normal samples}, but higher scores for abnormal images. Hence a higher anomaly score $\mathcal{A}(x)$ for a given sample $x$ would indicate whether $x$ is any abnormal with respect to the distribution of normal data learned by $f$ from ${D}_{trn}$ during training.

\subsubsection{Pipeline}
\begin{figure*} 
    \centering
        \includegraphics[width=\textwidth]{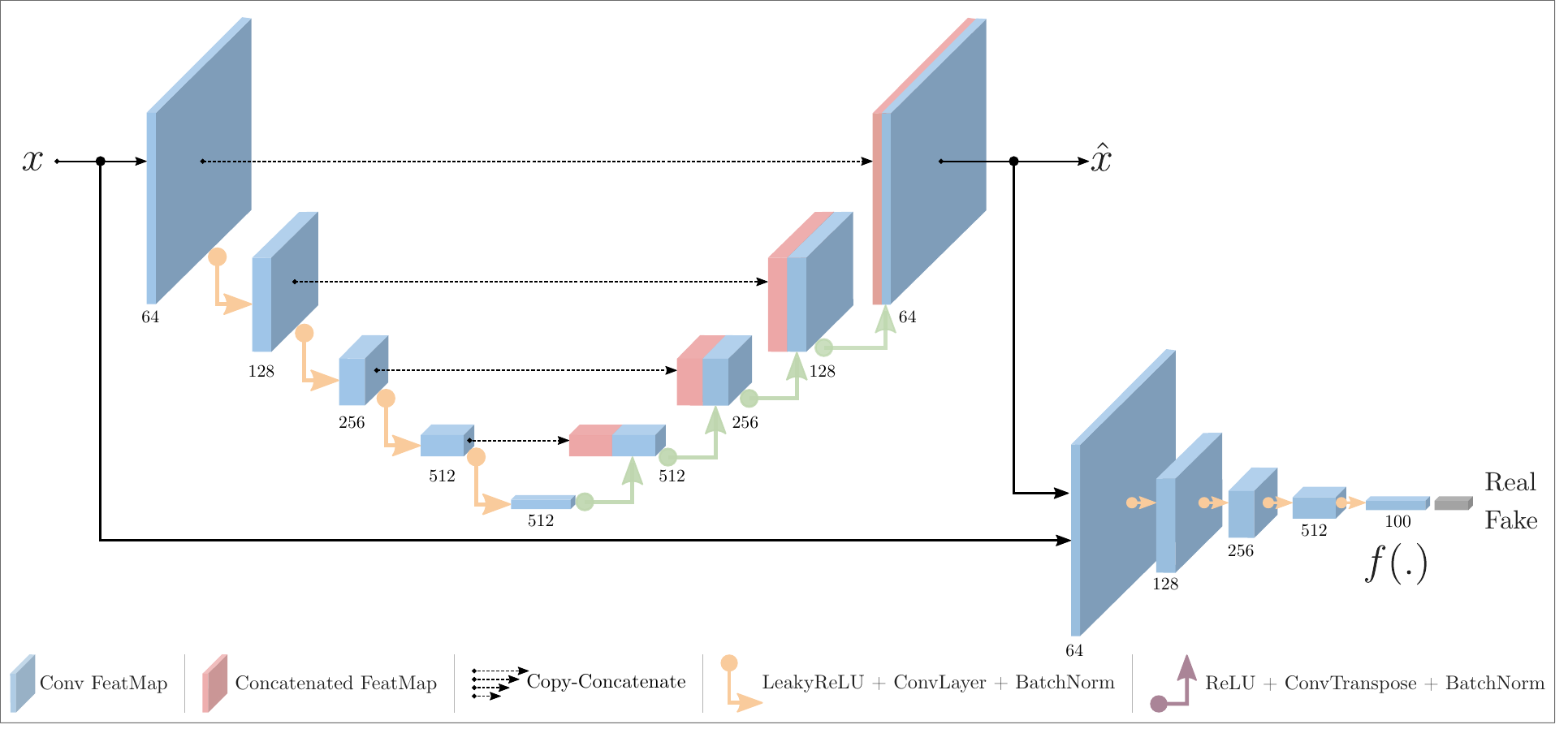}
    \caption{Details of the proposed network architecture.} 
    \label{fig:pipeline}
\end{figure*}

Figure \ref{fig:high-level-overview} shows a high-level overview of the proposed approach, which comprises a generator ($G$) and a discriminator ($D$) networks, respectively. The network $G$ adopts a bow-tie network using an encoder ($G_E$) and a decoder ($G_D$) networks. The encoder network captures the distribution of the input data by mapping high-dimensional image ($x$) into lower-dimensional latent representation ($z$) such that $G_E: x \rightarrow z$, where $x \in \mathbb{R}^{w \times h \times c}$ and $z \in \mathbb{R}^d$. As illustrated in Figure \ref{fig:pipeline}, the network $G_E$ reads input $x$ through five blocks containing Convolutional and BatchNorm layers as well as LeakyReLU activation function and outputs the latent representation $z$, which is also known as the bottleneck features that carries a unique representation of the input.

Being symmetrical to $G_E$, the decoder network $G_D$ up-samples the latent vector $z$ back to the input image dimension and reconstructs the output, denoted as $\hat{x}$. Motivated by \cite{ronneberger2015unet}, the decoder $G_D$  adopts skip-connection approach such that each down-sampling layer in the encoder network is concatenated to its corresponding up-sampling decoder layer (Figure \ref{fig:pipeline}). This use of skip connections provides substantial advantages via direct information transfer between the layers, preserving both local and global (multi-scale) information, and hence yielding better reconstruction.   

The second network within the pipeline, shown in Figure \ref{fig:pipeline} (b), called discriminator ($D$), predicts the class label of the given input. In this context, its task is to classify real images ($x$) from the fake ones ($\hat{x}$), generated by the network $G$. The network architecture of the discriminator $D$ follows the same structure as the discriminator of the DCGAN approach presented in \cite{Radford2015}. Besides being a classifier, the network $D$ is also used as a feature extractor such that latent representations of the input image $x$ and the reconstructed image $\hat{x}$ are computed. Extracting the features from the discriminator to perform inference within the latent space is the novel part of the proposed approach compared to the previous approaches \cite{Schlegl2017, Zenati2018a, Akcay2018b}. 

Based on this multi-network architecture, explained above and shown in Figure \ref{fig:pipeline}, the next section describes the proposed training objective and inference scheme.

\subsection{Training Objective}
As explained in Section \ref{sssec:problem-definition}, the idea proposed in this work is to train the model only on normal samples, and test on both normal and abnormal ones. The motivation is that we expect the model to be able to correctly reconstruct the normal samples either in image or latent vector space. The hypothesis is that the network is conversely expected to fail to reconstruct the abnormal samples as it is never trained on such abnormal examples. Hence, for abnormal samples, one would expect a higher loss for the reconstruction of the output image representation $\hat{x}$ or the latent representation $\hat{z}$. To validate this, we propose to combine three loss values (\textit{Adversarial, Contextual, Latent}), each of which has its own contribution to make within the overall training objective. 

\subsubsection{Adversarial Loss}
\label{sssec:adversarial-loss}
In order to maximize the reconstruction capability for the normal images $x$ during training, we utilize the adversarial loss proposed in \cite{Goodfellow2014b}. This loss, shown in Equation \ref{eq:adversarial-loss}, ensures that the network $G$ reconstructs a normal image $x$ to $\hat{x}$ as realistically as possible, while the discriminator network $D$ classifies the real and the (fake) generated samples. The task here is to minimize this objective for $G$, and maximize for $D$ to achieve $\underset{G}{\text{min }} \underset{D}{\text{max }} \mathcal{L}_{adv}$, where $\mathcal{L}_{adv}$ is denoted as

\begin{equation} \label{eq:adversarial-loss}
    \mathcal{L}_{adv} =  \underset{x \sim p_x}{\mathbb{E}} [\log D(x)] + \underset{x \sim p_x}{\mathbb{E}} [\log (1-D(\hat{x})].
\end{equation}

\subsubsection{Contextual Loss}
The adversarial loss defined in Section \ref{sssec:adversarial-loss} impose the model to generate realistic samples, but does not guarantee to learn contextual information regarding the input. To explicitly learn this contextual information to sufficiently capture the input data distribution for the normal samples, we apply $L_1$ normalization to the input $x$ and the reconstructed output $\hat{x}$. This normalization ensures that the model is capable of generating contextually similar images to normal samples. The contextual loss of the training objective is shown below:

\begin{equation} \label{eq:contextual-loss}
    \mathcal{L}_{con} = \underset{x \sim p_x}{\mathbb{E}} \lvert x - \hat{x} \rvert_1.
\end{equation}

\subsubsection{Latent Loss} 
With the adversarial and contextual losses defined above, the model is able to generate realistic and contextually similar images. In addition to these objectives, we aim to reconstruct latent representations for the input $x$ and the generated normal samples $\hat{x}$ as similar as possible. This is to ensure that the network is capable of producing contextually sound latent representations for common examples. As depicted in Figure \ref{fig:pipeline}(b), we use the final convolutional layer of the discriminator $D$, and extract the features of $x$ and $\hat{x}$ to reconstruct their latent representations such that $z = f(x)$ and $\hat{z} = f(\hat{x})$. The latent representation loss therefore becomes:

\begin{equation} \label{eq:latent-representation-loss}
    \mathcal{L}_{lat} = \underset{x \sim p_x}{\mathbb{E}} \lvert f(x) - f(\hat{x}) \rvert_2.
\end{equation}

Finally, total training objective becomes a weighted sum of the losses above.
\begin{equation} \label{eq:training-objective}
    \mathcal{L} = \lambda_{adv}\mathcal{L}_{adv} + 
                  \lambda_{con}\mathcal{L}_{con} + 
                  \lambda_{lat}\mathcal{L}_{lat}
\end{equation}
where $\lambda_{adv}$, $\lambda_{con}$ and $\lambda_{lat}$ are the weighting parameters adjusting the dominance of the individual losses to the overall objective function.

\subsection{Inference}
To find the anomalies during the testing and subsequent deployment, we adopt the anomaly score, proposed in \cite{Schlegl2017} and also employed in \cite{Zenati2018a}. For a given test image $\dot{x}$, its anomaly score becomes:

\begin{equation} \label{eq:anomaly-score}
    \mathcal{A}(\dot{x}) = \lambda R(\dot{x})  + (1 - \lambda) L(\dot{x})
\end{equation}
where $R(\dot{x})$ is the reconstruction score measuring the contextual similarity between the input and the generated images based on Equation \ref{eq:contextual-loss}. $L(\dot{x})$ denotes the latent representation score measuring the difference between the input and generated images based on Equation \ref{eq:latent-representation-loss}. $\lambda$ is the weighting parameter controlling the relative importance of the score functions. 

Based on Equation \ref{eq:anomaly-score}, we then compute the anomaly scores for each individual test sample $\dot{x}$ in the test set $\mathcal{D}_{tst}$, and denote as anomaly score vector $\boldsymbol{A}$ such that $\boldsymbol{A} = \{A_i: \mathcal{A}(\dot{x}_i), \dot{x}_i \in \mathcal{D}_{tst}\}$. Finally, following the same procedure proposed in \cite{Akcay2018b}, we also apply feature scaling to $\boldsymbol{A}$ to scale the anomaly scores within the probabilistic range of $[0, 1]$. Hence, the updated anomaly score for an individual test sample $\dot{x}$ becomes:

\begin{equation} \label{eq:anomaly-score-scaled}
    \hat{\mathcal{A}}(\dot{x})  = \frac{\mathcal{A}(\dot{x}) - min(\boldsymbol{A})}{max(\boldsymbol{A}) - min(\boldsymbol{A})}.
\end{equation}

Equation \ref{eq:anomaly-score-scaled} finally yields an anomaly score vector $\hat{\boldsymbol{A}}$ for the final evaluation of the test set $\mathcal{D}_{tst}$, which is explained in Sections \ref{ssec:evaluation} and \ref{sec:results}.
    \section{Experimental Setup}
\label{sec:experimental-setup}
This section introduces the datasets, training and implementational details as well as the evaluation criteria used within the experimentation.

\begin{table*}[t]
    \centering
    \begin{tabular}{@{}lllllllllll@{}}
    \toprule
             & \multicolumn{10}{c}{CIFAR-10}                                                                                                                                           \\ \cmidrule(l){2-11} 
    Model    & bird           & car            & cat            & deer           & dog            & frog           & horse          & plane          & ship           & truck          \\ \midrule
    AnoGAN \cite{Schlegl2017}   & 0.411          & 0.492          & 0.399          & 0.335          & 0.393          & 0.321          & 0.399          & 0.516          & 0.567          & 0.511          \\
    EGBAD \cite{Zenati2018a}    & 0.383          & 0.514          & 0.448          & 0.374          & 0.481          & 0.353          & 0.526          & 0.577          & 0.413          & 0.555          \\
    GANomaly \cite{Akcay2018b} & \textbf{0.510} & 0.631          & 0.587          & 0.593          & \textbf{0.628} & 0.683          & 0.605          & 0.633          & 0.616          & 0.617          \\
    \textbf{Proposed} & 0.448          & \textbf{0.953} & \textbf{0.607} & \textbf{0.602} & 0.615          & \textbf{0.931} & \textbf{0.788} & \textbf{0.797} & \textbf{0.659} & \textbf{0.907} \\ \bottomrule
    \end{tabular}
    \caption{AUC results for CIFAR-10 dataset}
    \label{tab:cifar-results}
\end{table*}

    \subsection{Datasets}
    To demonstrate the proof of concept of the proposed approach, we validate the model on four different datasets, each of which is explained in the following subsections.
    
    We perform our evaluation using the benchmark CIFAR-10 dataset \cite{cifar10} and also the UBA and FFOB datasets \cite{Akcay2018b}. Using CIFAR-10 we formulate a \textit{leave one class out} anomaly detection problem. For the application context of X-ray baggage screening \cite{Akcay2018}, the UBA and FFOB datasets from \cite{Akcay2018b} are used to formulate an anomaly detection problem based on the concept of weapon threat items being an anomaly within the security screening process.

        
        \subsubsection{CIFAR-10}
        Experiments for the CIFAR-10 dataset has the one versus the rest approach. Following this procedure yields ten different anomaly cases for CIFAR-10, each of which has $45,000$ normal training samples, and  $9,000$:$6,000$ normal-abnormal test samples.

        \subsubsection{University Baggage Dataset ---UBA}
        This in-house dataset comprises 230,275 dual energy X-ray security image patches extracted via a $64 \times 64$ overlapping sliding window approach. The dataset contains 3 abnormal sub-classes ---\textit{knife (63,496), gun (45,855) and gun component (13,452)}. Normal class comprises 107,472  benign X-ray patches, splitted via 80:20 train-test ratio. 

        \subsubsection{Full Firearm vs Operational Benign ---FFOB}
        As presented in \cite{Akcay2018b}, we also evaluate the performance of the model on the UK government evaluation dataset \cite{CAST2016}, comprising both expertly concealed firearm (threat) items and operational benign (non-threat) imagery from commercial X-ray security screening operations (baggage/parcels). Denoted as FFOB, this dataset comprises 4,680 firearm full-weapons as full abnormal and 67,672 operational benign as full normal images, respectively.         

    \subsection{Training Details}
\begin{figure}
    \centering
    \includegraphics[width=\linewidth]{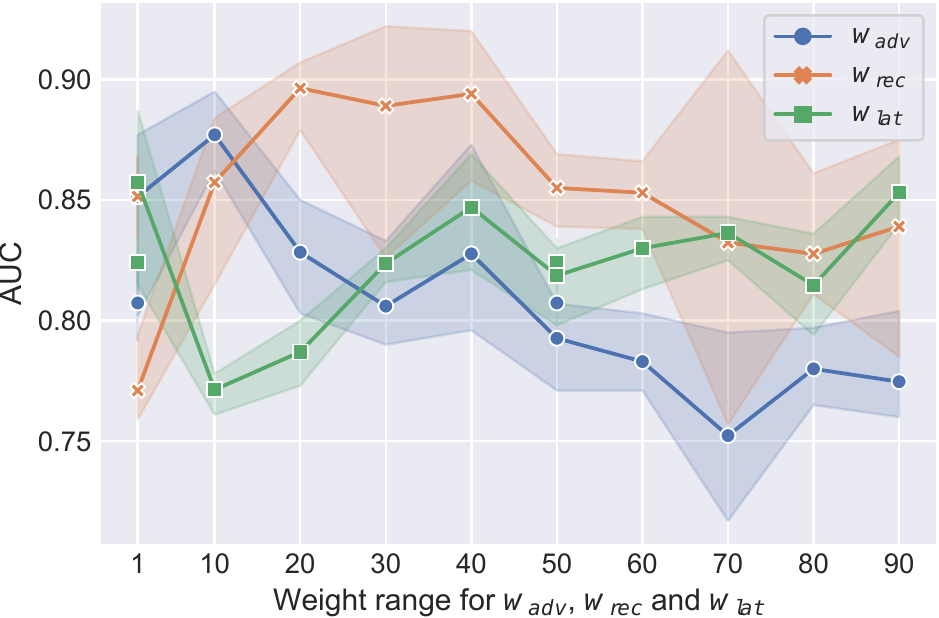}
    \caption{Hyper-parameter tuning for the model. The model achieves the most optimum performance when $\lambda_{adv} = 1$, $\lambda_{rec=40} = 1$ and $\lambda_{enc} = 1$. }
    \label{fig:hyper-parameter-weights}
\end{figure}

    The training objective $\mathcal{L}$ from Equation \ref{eq:training-objective} is optimized via Adam\cite{Kingma2014} optimizer with an initial learning rate $lr=2e^{-3}$ with a lambda decay, and momentums $\beta_1=0.5$, $\beta_2=0.999$. The weighting parameters of $\mathcal{L}$ is chosen as $\lambda_{adv}=1$, $\lambda_{rec}=40$ and $\lambda_{lat}=1$, empirically shown to yield the optimal performance (See Figure \ref{fig:hyper-parameter-weights}). The model is initially set to be trained for 15 epochs; however, in most cases it learns sufficient information within less training cycles. Therefore, we save the parameters of the network when the performance of the model starts to decrease since this reduce is a strong indication of over-fitting. The model is implemented using PyTorch \cite{Paszke2017} (v0.5.1, Python 3.7.1, CUDA 9.3 and CUDNN 7.1). Experiments are performed using an NVIDIA Titan X GPU. 

    \subsection{Evaluation}
    \label{ssec:evaluation}
    The performance of the model is evaluated by the area under the curve (AUC) of the receiver operating characteristics (ROC) \cite{AUC}, a function plotted by the true positive rates (TPR) and false positive rates (FPR) with varying threshold values (as per prior work in the field \cite{Schlegl2017,Zenati2018a,Akcay2018b}

    \section{Results}
\label{sec:results}
\begin{figure}
    \centering
    \includegraphics[width=\linewidth]{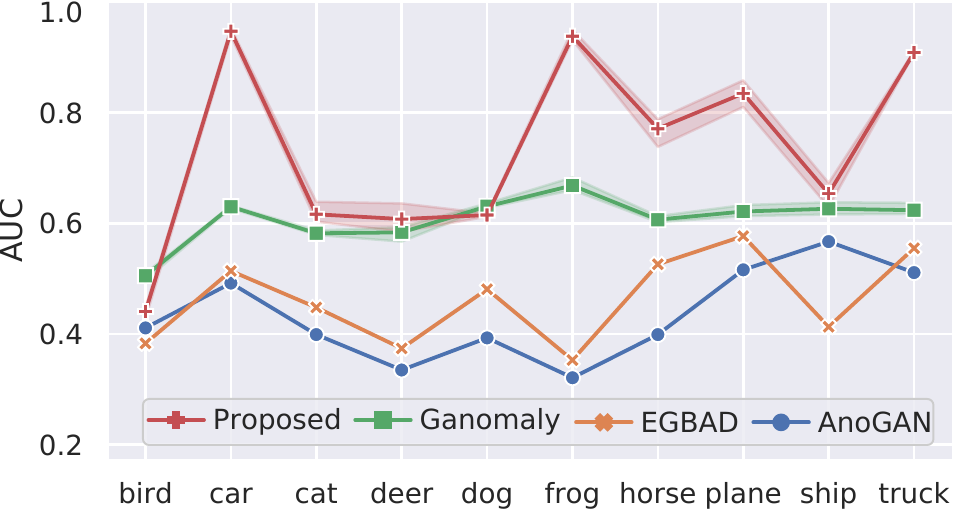}
    \caption{AUC results for CIFAR-10 dataset. Shaded areas in the plot represents variations due to the use of 3 random seeds.}
    \label{fig:cifar-results}
\end{figure}

For the CIFAR-10 dataset, Table \ref{tab:cifar-results} / Figure \ref{fig:cifar-results} demonstrate with the exception of abnormal classes \textit{bird} and \textit{dog}, the proposed model yield superior results to the prior work.  

Table \ref{tab:xray-results} presents the experimental results for UBA and FFOB datasets. It is apparent from this table that the proposed method significantly outperforms the prior work in each anomaly cases of the datasets. Of significance, the best AUC of the prior work is $0.599$ for the most challenging abnormality case -- \textit{knife}, while the method proposed here achieves AUC of  $0.904$. 

\begin{table}[h]
    \centering
    \begin{tabular}{@{}lcccccc@{}}  \toprule
                & \multicolumn{4}{c}{UBA}                                           &  & \multicolumn{1}{c}{FFOB} \\ \cmidrule(lr){2-5} \cmidrule(l){7-7}
    Method   & gun            & gun-parts      & knife          & overall        &  & full-weapon              \\ \midrule
    AnoGAN \cite{Schlegl2017}  & 0.598          & 0.511          & {0.599} & 0.569          &  & 0.703                    \\
    EGBAD \cite{Zenati2018a}  & 0.614          & 0.591          & 0.587          & 0.597          &  & 0.712                    \\
    GANomaly \cite{Akcay2018b} & {0.747} & {0.662} & 0.520          & {0.643} &  & {0.882}    \\ 

    \textbf{Proposed} & \textbf{0.972} & \textbf{0.945} & \textbf{0.904}         & \textbf{0.940} &  & \textbf{0.903}    \\ 

    \bottomrule
    \end{tabular}
    \caption{AUC results for UBA and FFOB datasets}
    \label{tab:xray-results}
\end{table}

\begin{figure*}
    \centering
    \includegraphics[width=\linewidth]{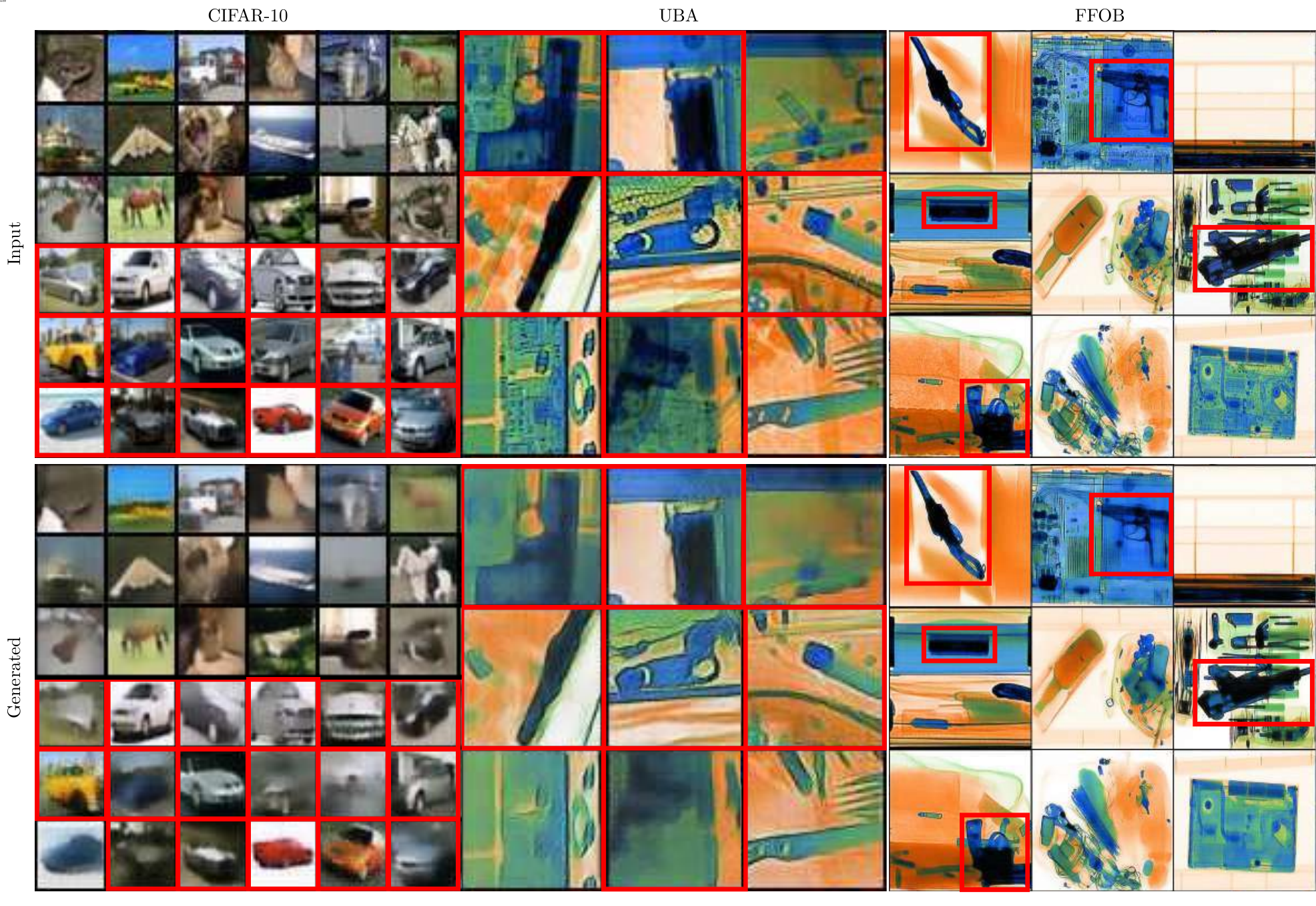}
    \caption{Exemplar test images for CIFAR-10, UBA and FFOB datasets when the abnormalities are car, gun-gun component-knife and gun, respectively. Despite the model's capability of generating even abnormal samples, the proposed model is able to detect abnormality within latent object space.}
    \label{fig:results}
\end{figure*}

Figure \ref{fig:results} depicts exemplar test images for the datasets used in the experimentation. A significant result emerging from the examples presented within Figure \ref{fig:results} is that the proposed model is capable of generating both normal and abnormal reconstructed outputs at test time, meaning that it captures the distribution of both domains. This is probably due to the use of skip connections enabling reconstruction even for the abnormal test samples.

The qualitative results of Figure \ref{fig:results}, supporting by the quantitative results of Table \ref{tab:xray-results} reveal that abnormality detection is successfully made in latent object space of the model that emerges from our adversarial training over the proposed skip-connected architecture. 

\begin{figure}
    \centering
    \includegraphics[width=\linewidth]{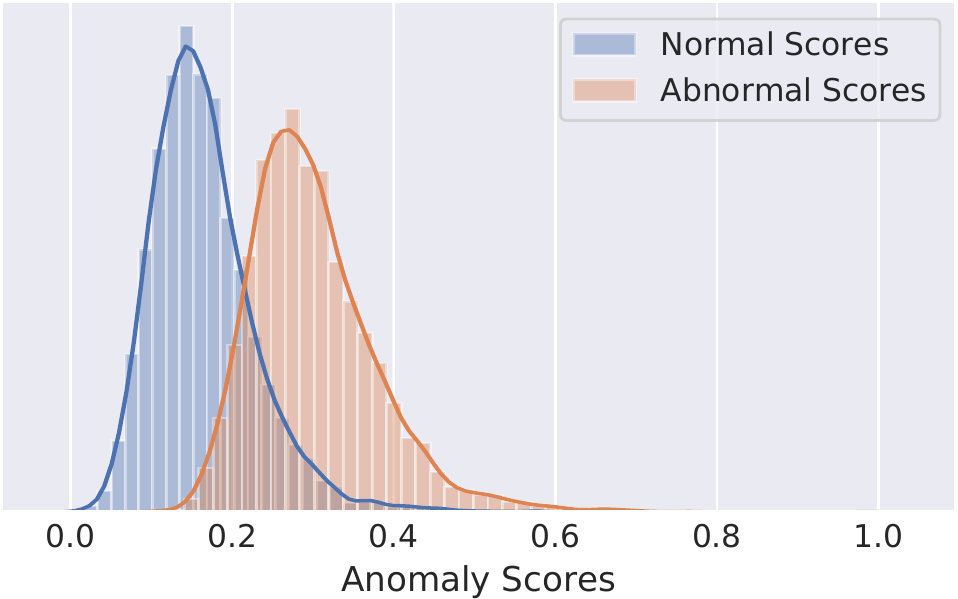}
    \caption{(a) Histogram of the normal and abnormal scores for the test data.}
    \label{fig:histogram}
\end{figure}

\begin{figure}
    \centering
    \includegraphics{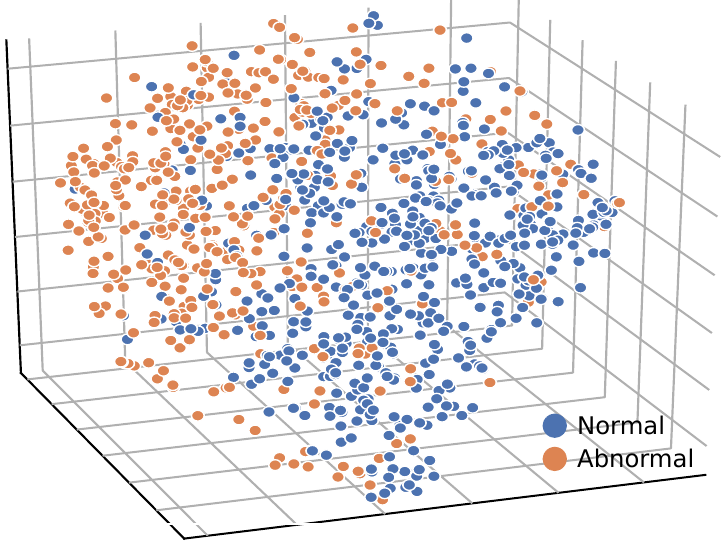}
    \caption{(b) t-SNE plot of the 1000 subsampled normal and abnormal features extracted from  the last convolutional layer ($f(.)$) of the discriminator (Figure \ref{fig:pipeline}).}
    \label{fig:tsne}
\end{figure}

Figures \ref{fig:histogram} and \ref{fig:tsne} show the  histogram plot (a) of the normal and abnormal scores for the test data, and the t-SNE plot (b) of the normal and abnormal features extracted from  the last convolutional layer ($f(.)$) of the discriminator (see Figure \ref{fig:pipeline}). Closer inspection of the figures reveals that the model yields promising separation within both the output anomaly (reconstruction) score and the preceding convolutional feature spaces.

Overall, these results indicate that the proposed approach yields superior anomaly detection performance to the previous \sota approaches. 

\begin{figure}
    \centering
    \includegraphics[width=\linewidth]{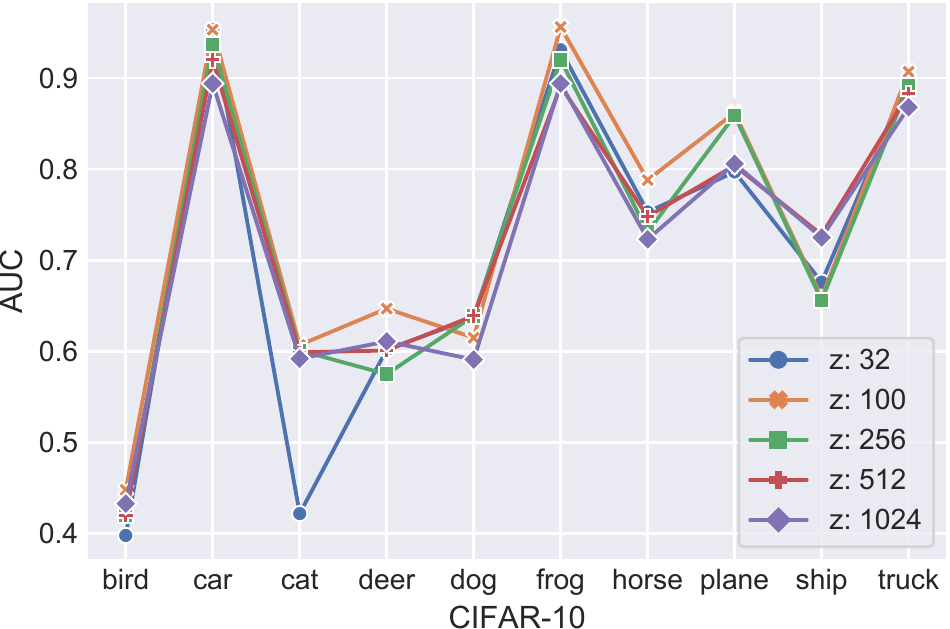}
    \caption{Hyper-parameter tuning for the model. The model achieves the most optimum performance when $nz=100$. }
    \label{fig:hyper-parameter-nz}
\end{figure}

    \section{Conclusion}
\label{sec:conclusion}
This paper introduces a novel unsupervised anomaly detection architecture within an adversarial training scheme. The proposed approach examines the role of skip connections within the generator and feature extraction from the discriminator for the manipulation of hidden features. Based on an evaluation across multiple datasets from different domains and complexity, the findings indicate that skip connections provide more stable training, and the inference learning from the discriminator achieves numerically superior results than the previous \sota methods. The empirical findings in this study provide an insight into the generalization capability of the proposed method to any anomaly detection task. Further research could also be conducted to determine the effectiveness of the proposed approach on both higher resolution images and various other anomaly detection tasks containing temporal information.

{
    \small
    \bibliographystyle{IEEEtran}
    \bibliography{ref/library}
}

    \clearpage
\end{document}